\documentclass[conference]{IEEEtran}
\usepackage{soul}
\usepackage{url}
\usepackage{mathtools}
\usepackage{pgfplots}
\usepackage{graphicx}
\usepackage{dblfloatfix} 
\usepackage{float}
\usepackage[T1]{fontenc}
\usepackage{amssymb}
\usepackage[caption=false]{subfig}
\usepackage{pdfpages}
\usepackage{wrapfig}
\usepackage{booktabs}
\usepackage{cite}

\newcommand{\model}{$\beta$-VAEGAN } 
\newcommand{\nmodel}{$\beta$-VAEGAN} 
\newcommand{\D}{\mathcal{D}}
\newcommand{\G}{\mathcal{G}}
\newcommand{\tD}{$\mathcal{D}$ }
\newcommand{\tG}{$\mathcal{G}$ }
\newcommand{\secondDataset}{Wafer }  
\AtBeginDocument{%
  \providecommand\BibTeX{{%
    \normalfont B\kern-0.5em{\scshape i\kern-0.25em b}\kern-0.8em\TeX}}}

\begin{document}
\title{Adversarial Anomaly Detection using Gaussian Priors and Nonlinear Anomaly Scores}

\author{\IEEEauthorblockN{Fiete Lüer}
\IEEEauthorblockA{\textit{eMundo Gmbh} \\
\textit{Gofore Oyj} \\
Munich, Germany \\
fiete.lueer@e-mundo.de}
\and
\IEEEauthorblockN{Tobias Weber}
\IEEEauthorblockA{\textit{Department of Statistics} \\
\textit{LMU Munich}\\
Munich, Germany \\
tobias.weber@stat.uni-muenchen.de}
\and
\IEEEauthorblockN{Maxim Dolgich}
\IEEEauthorblockA{\textit{eMundo GmbH} \\
\textit{Gofore Oyj}\\
Munich, Germany \\
maxim.dolgich@e-mundo.de}
\and
\IEEEauthorblockN{Christian Böhm}
\IEEEauthorblockA{\textit{Faculty of Computer Science} \\
\textit{University of Vienna}\\
Vienna, Austria \\
christian.boehm@univie.ac.at}}

\maketitle
\makeatletter
 \let\old@ps@headings\ps@headings
 \let\old@ps@IEEEtitlepagestyle\ps@IEEEtitlepagestyle
 \def\confheader#1{%
 \def\ps@headings{%
 \old@ps@headings%
 \def\@oddhead{\strut\hfill#1\hfill\strut}%
 \def\@evenhead{\strut\hfill#1\hfill\strut}%
 }%
 \def\ps@IEEEtitlepagestyle{%
 \old@ps@IEEEtitlepagestyle%
 \def\@oddhead{\strut\hfill#1\hfill\strut}%
 \def\@evenhead{\strut\hfill#1\hfill\strut}%
 }%
 \ps@headings%
 }
 \makeatother

\confheader{%
 Accepted at 2023 IEEE International Conference on Data Mining Workshops (ICDMW).
 }

\begin{abstract}
Anomaly detection in imbalanced datasets is a frequent and crucial problem, especially in the medical domain where retrieving and labeling irregularities is often expensive. 
By combining the generative stability of a $\beta$-variational autoencoder (VAE) with the discriminative strengths of generative adversarial networks (GANs), we propose a novel model, \nmodel.
We investigate methods for composing anomaly scores based on the discriminative and reconstructive capabilities of our model. Existing work focuses on linear combinations of these components to determine if data is anomalous. We advance existing work by training a kernelized support vector machine (SVM) on the respective error components to also consider nonlinear relationships. This improves anomaly detection performance, while allowing faster optimization. Lastly, we use the deviations from the Gaussian prior of \model to form a novel anomaly score component.
In comparison to state-of-the-art work, we improve the $F_1$ score during anomaly detection from 0.85 to 0.92 on the widely used MITBIH Arrhythmia Database.\footnote{\label{note1}The source code is available at \url{https://github.com/emundo/ecgan}.}
\end{abstract}


\section{Introduction}
Anomaly detection (AD) commonly needs to be performed on highly imbalanced datasets with applications ranging from industrial sensor data to the detection of pathologies in medical data. Especially in medical applications, not all known abnormal classes are sufficiently represented in datasets and some pathologies are likely currently unknown. 
To reduce the risk of omitting relevant classes, the classification is frequently investigated in a one-class (OC) setting:
\textit{How can we detect anomalies if only the normal class is available during training?} \\
%
%
\begin{figure}[!t]
\centering
\includegraphics[width=0.5\textwidth, trim=0 0 50 0, clip]{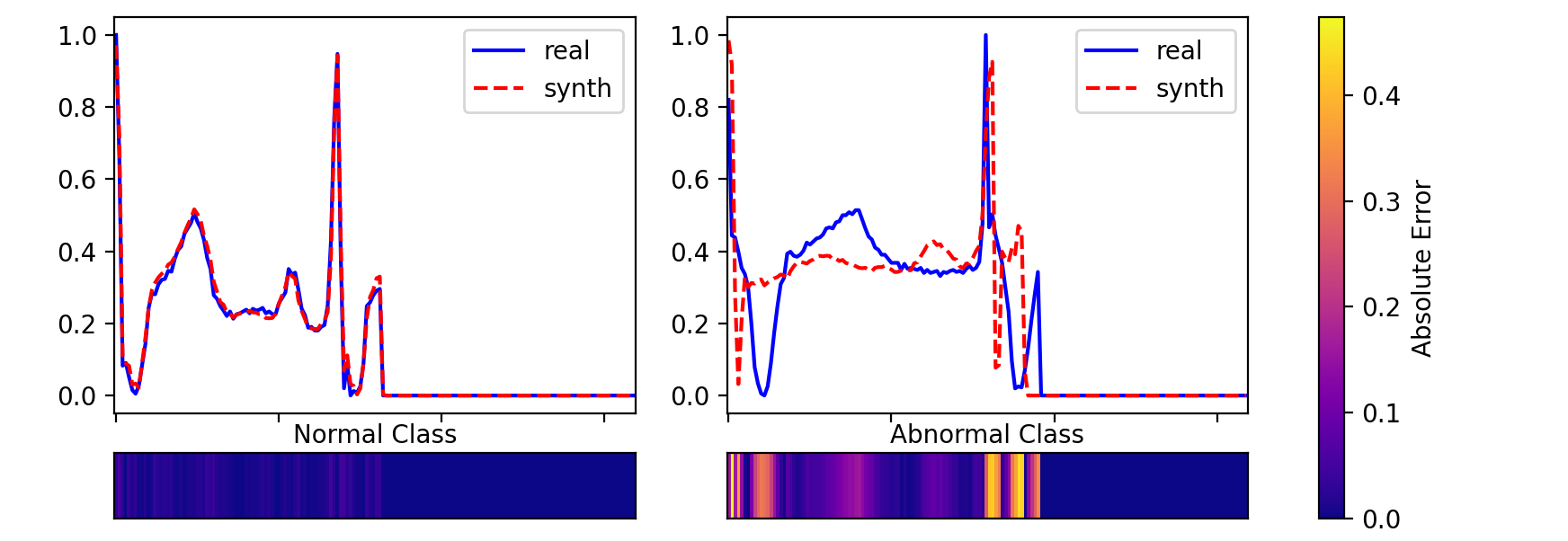}
\caption{Heatmap of the absolute error between real ECG samples and synthetic reconstructions of the normal (left) and abnormal (right) class.}
\label{fig:reconstructed sample}
\end{figure}
%
%
In the OC setting, generative deep learning models such as GANs \cite{gan} have frequently been applied and improved in recent years. Schlegl et al. \cite{anogan} use the reconstructive as well as discriminative capabilities of GANs to detect anomalies, but the proposed procedure does not allow real-time monitoring due to high computational costs during inference. Zhou et al. \cite{beatgan} introduce BeatGAN to circumvent this problem by using an encoder to learn an embedding. One major drawback of this approach is the ability of autoencoders (AEs) to generate out-of-distribution (OOD) data \cite{NAE}. 
In this work, we improve the generative capabilities of BeatGAN by using recent advances in adversarial training. We propose \nmodel, an adversarially regularized $\beta$-VAE with a Gaussian latent prior to enforce in-distribution generation and embeddings. This is later used to improve the discrimination of abnormal samples. We investigate existing anomaly score compositions, currently dominated by linear combinations, and leverage nonlinear dependencies of the  components to improve AD performance. \\ 
We compare \model to existing state-of-the-art methods on the public arrhythmia dataset MITBIH \cite{mitbih} and validate it on an industrial silicon wafer dataset \cite{wafer}.
Our contributions to existing research lie in
\begin{itemize}
\item \textbf{Improving the quality of generated data:} We improve the quality of generated data, increasing the normalized TSTR score from 0.90 to 0.95 in comparison to state-of-the-art work.
\item \textbf{Using latent space information for anomaly detection:} We use properties of the $\chi$ distribution present in standard Gaussian distributed latent spaces, introducing a novel anomaly score component and discussing further uses of the latent norms.
\item \textbf{Leveraging nonlinear anomaly scores:} Using an SVM with a nonlinear RBF kernel, we improve speed as well as performance of the AD procedure.
\end{itemize}
Fully reproducible implementations and results have been published as part of a novel framework for anomaly detection.
\section{Related Work}
One-class settings frequently occur in the medical domain, but traditional approaches such as OCSVMs and isolation forests often do not yield sufficient results. More recently, neural network based models have been researched extensively. These approaches often utilize a thresholded reconstruction error, where usually a generative architecture is trained only on the distribution of normal data and tasked to reconstruct a given data sample. A basic assumption is that anomalous data could only be generated with limited resemblance to the original. 
Another category of approaches concerns with maximizing the (log-)likelihood of regular data, using the output score of a network to classify the data (e.g. \cite{ecgrcnn}). These approaches often lack interpretability, are easily overconfident in their predictions \cite{calibration}, and handling high dimensional data is difficult \cite{waic}.
\paragraph{Generative Adversarial Networks}
A GAN consists of two networks, a generator \tG and discriminator $\D$. The generative function $\mathcal{G}: z\mapsto x$ maps a low dimensional latent vector $z\in \mathbb{R}^k$ sampled from a predefined distribution $p_z$ to a high dimensional sample $x\in \mathcal{X}$ where $\mathcal{X}$ is the input space. The discriminator $\D$ maps real samples $x$ and synthetic samples $\G(z)$ to scalars, indicating the confidence that they are sampled from the training distribution $p_x \in \mathcal{X}$.
Both components can be used to detect anomalies and various GAN-based AD models solely utilize either - the discrimination error (e.g. \cite{disconly}) or the reconstruction error (e.g. \cite{beatgan}, \cite{multiple_latent_starts}) during inference. To reduce the previously mentioned negative impact of the direct discriminator output, AnoGAN \cite{anogan} utilizes both components to assign a linearly weighted anomaly score consisting of a reconstruction score $R(x)$ and discrimination score $D(x)$.\\
To calculate $R(x)$ and $D(x)$, AnoGAN requires an arbitrarily ($\epsilon$-) similar generated point $\G(\hat{z})$ for a given inference sample $x$. $\hat{z}$ is retrieved by a latent optimization scheme.
The resulting $R(x)$ is the reconstruction loss $\mathcal{L}_R(x, \hat{z})$, often the mean squared error (MSE) between $x$ and $\G(\hat{z})$. Since using the output of classification networks is faulty for high dimensional data \cite{waic}, $D(x)$ corresponds to the feature difference of the discriminator to form the discriminative loss $\mathcal{L}_\D$. This difference is retrieved by comparing the feature activations $f(x)$ of all layers up to the penultimate layer, forming the feature matching loss (\cite{improved_techniques}, \cite{anogan}): $\mathcal{L}_\D(x,z)=\sum |f(x)-f(\G(z))|$.
This suffers from two major drawbacks. First, stabilizing GAN training is difficult, frequently suffering from non-convergence or collapsing towards one or few modes \cite{improved_techniques}.
Second, the non-convex optimization problem introduces an additional source of error, which can be slightly reduced by selecting multiple starting points $z_i$ of the optimization process \cite{multiple_latent_starts}, and leads to a detrimental problem for many practical applications: Optimization takes several seconds for a single datum, which renders real-time inference unfeasible. 
More recent work has proposed the addition of a mapping from input to latent space to capture the inversion process explicitly (e.g. \cite{alad}, \cite{fanogan}).
\paragraph{Adversarial Autoencoders}
To avoid the training instabilities, \tG can receive explicit feedback using $x$ as its input. 
Previous advances in adversarial AEs include Larsen et al. \cite{VAEGAN} who introduce VAE-GAN to combine the training stability of (V)AEs with the high fidelity of GANs using the discriminators' adversarial error. While VAE-GAN focuses on data generation, subsequent work introduces AnoVAEGAN \cite{anovaegan} to detect anomalies in medical imaging, also considering information on the latent space by measuring the KL-divergence between the multivariate normal from which data is drawn and the distribution of all generated z. Zhou et al. \cite{beatgan} propose a current state-of-the-art model for medical time series generation, BeatGAN, where the reconstruction error of an adversarial AE is considered to detect abnormalities in cardiac data. The encoder of these AE-based approaches can be leveraged to achieve the inverse mapping required for faster inference.
We expand existing work by incorporating the $\beta$-VAE framework to control disentanglement and assimilation to the prior distribution where the prior information is utilized in a novel way. We further add spectral normalization and extend the anomaly score components and combination.
\section{Methods}
GANs describe a two-player min-max game between a discriminator \tD and generator \tG trained using a joint value function $V$:
\begin{equation}
\label{eq:gan}
    \min_{\G} \max_{\D} V(\D,\G)=\text{ }
    \mathbb{E}_{x\sim p_x}
    [\text{log } \D(\textit{x})]
    +\mathbb{E}_{z\sim p_z}
    [\text{log}(1-\D(\G(\textit{z})))],
\end{equation}
where $p_z$ usually is $\mathcal{N}(0, I)$.
In contrast to GANs, AEs use $x$ to encode a low dimensional representation $enc_\phi(x)=z$ resulting in a reconstructed sample $\hat{x}$ retrieved from decoder $dec_\theta(z)= \hat{x}$, parameterized by $\phi$ and $\theta$ respectively.
The objective of AEs is to solely minimize the reconstruction error resulting in a non-stationary latent distribution, which does not follow a predefined distribution.\\
Variational autoencoders \cite{vae} assume a prior distribution $p(z)$ 
and optimize two objectives: (i) maximizing the log-likelihood of generating real data, (ii) minimizing the $\mathbb{KL}$ divergence between the prior and the posterior distribution.
In a VAE $enc_\phi$ approximates this posterior inference distribution $q_\phi(z|x)=\mathcal{N}(\mu_\phi,\text{diag}(\sigma^2_\phi))$ by estimating $\mu_\phi$ and $\sigma_\phi$. 
Afterwards $z\sim q_\phi(z|x)$ is sampled to reconstruct $x$ using the decoder $dec_\theta=p_\theta(x|z)$. In the following, the output of $dec_\theta(z)$ is also described as $\mathcal{G}(z)$. 
The reconstruction is optimized using the negative evidence lower bound (ELBO), which can be extended using a weighting factor $\beta$ \cite{betavae}
\begin{equation}
\mathcal{L}_{VAE}=-\text{log } p_\theta (x|z)+\beta \cdot \textstyle \mathbb{KL} (q_\phi(z|x) \Vert p(z)),
\end{equation}
resulting in the original VAE objective if $\beta=1$.\\
The AnoGAN interpolation can now be substituted by the encoder network of the (V)AE, allowing the retrieval of an inverse sample $\hat{z}$ using a single forward pass through $enc_\phi$.
The resulting output of the decoder $\G(\hat{z})$ can be used to calculate the reconstruction score $R(x)=\text{MSE}(x,\G(\hat{z}))=\text{MSE}(x,\hat{x})$. 
Similarly, we use the MSE to calculate the feature matching loss  $D(x)=\text{MSE}(f(x),f(\G(\hat{z}))=\text{MSE}(f(x),f(\hat{x}))$ resulting in a linearly weighted anomaly score  $A(x) = (1-\lambda) \cdot R(x) + \lambda \cdot D(x)$ (from \cite{anogan}).
Data is labeled as anomalous if $A(x)$ exceeds a threshold $\tau$.
\subsection{\nmodel}
In \nmodel, the deterministic AE used in BeatGAN is replaced by the probabilistic $\beta$-VAE. 
The discriminator still minimizes the binary cross entropy (BCE) trained on real data $x$ and reconstructed data $\G (z)=dec_\theta(z)=\hat{x}$
\begin{equation}
    \mathcal{L}_{\D} =\text{log }\D (x) + \text{log }(1- \D (\G (z))).
\end{equation}
The generator jointly optimizes $enc_\phi$ and $dec_\theta$, minimizing
\begin{equation}
\label{eq:gen_loss}
\resizebox{.91\linewidth}{!}{
\begin{math}
  \mathcal{L}_{\G} ={\underbrace{\textstyle \vert\vert f(x)-f(\G(z)\vert\vert_2^2}_{\mathclap{\mathcal{L}_{adv}}}}
    +\nu [
    {\underbrace{\textstyle -\text{log } p_\theta (x|z)}_{\mathclap{\mathcal{L}_{rec}}}} +
    \beta \cdot 
    {\underbrace{\textstyle \mathbb{KL} (q_\phi(z|x) \Vert p(z)}_{\mathclap{\mathcal{L}_{\mathbb{KL}}}}}].
\end{math}
}
\end{equation}
The adversarial feature matching loss $\mathcal{L}_{adv}$ is used to regularize $\mathcal{L}_{VAE}(=\mathcal{L}_{rec}+\beta\cdot\mathcal{L}_{\mathbb{KL}})$ controlled by weighting parameter $\nu$. $\mathcal{L}_{rec}$ is used to measure the pointwise difference between real and generated data measured by the binary cross entropy. 
$\beta$ allows a trade-off between optimal reconstructions ($\beta=0$), resulting in the AE objective function, and the complete assimilation of the latent space to 
$p(z)$ by minimizing the $\mathbb{KL}$ divergence ($\beta\mapsto \infty$). The selection of $\beta>1$ is additionally important to learn disentangled features \cite{betavae}, which is especially desirable in medical settings (e.g. \cite{disentangled_cardio}). 
We apply spectral normalization \cite{spectral_norm} to the discriminator,
which helps to achieve a more concise representation, avoids large weights, and leads to a more robust AD feature matching loss.
\begin{figure}[!b]
\centering
\hspace*{-0.4cm}
\includegraphics[width=0.55\textwidth]{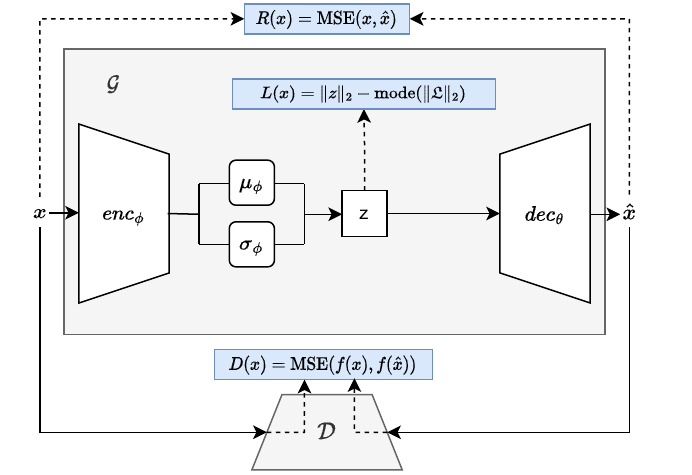}
\caption{Architecture of \nmodel. Blue components are anomaly detection losses.}
\label{fig:architecture}
\end{figure}
\subsection{Anomaly Detection}
\label{section:ad}
The latent space of AEs is not inherently restricted and in a VAE with a Gaussian prior, $q_\phi(z|x)$ converges towards $\mathcal{N}(0,\text{I})$ with an increasing $\beta$.
As $p(z)$ has full support on the latent space, it is possible that even highly abnormal data can be reconstructed by the generative network.
In practice, the impact is not severe because \tG does not have sufficient signal to learn a meaningful representation and gradients far from the mean of the Gaussian distribution. 

\paragraph{$\chi$ Distribution}

The reconstruction of the AnoGAN latent optimization is usually restricted to a predefined amount of reconstruction steps, to reduce the computational time in case the reconstruction is stuck in local optima. It is nonetheless important to avoid faulty reconstructions, especially in medical settings.\\
We can use the structure of the latent space to avoid entering unlikely regimes of the latent space. Given a standard Gaussian distributed latent space, Choi et al. \cite{waic} argue that values with a high likelihood are usually OOD samples in high dimensional space: The resulting norm of the latent vectors of high likelihood samples is zero, while
in-distribution data is distributed in an annulus around $\sqrt{dim(z)}$. 
The norm of the latent vectors of a standard AE does not necessarily allow us to draw conclusions on their (ab)normality. While the reconstructions of VAEs are usually of a lower quality than normal AEs due to the trade-off with $\mathcal{L}_{\mathbb{KL}}$, we can use a standard Gaussian prior to investigate deviations from the latent norm. 
While $\sqrt{dim(z)}$ is a sufficient estimator for the annulus for high dimensional spaces, the distribution of the latent vectors can be investigated in more detail: If $z \sim \mathcal{N}(0,\text{I})$, the resulting norm of $z$ is $\chi$-distributed by definition. 
In this case, one can use the distance from the resulting mode of the $\chi$ distribution, $\sqrt{dim(z)-1}\text{ for } dim(z) \geq 1$, as a direct anomaly indicator to detect abnormal and unwanted reconstructions.
\begin{table*}[!hb]
\centering
\caption{MITBIH TSTR scores across five folds using a running mean window of 5.}
\resizebox{\linewidth}{!}{%
\addtolength\tabcolsep{3pt}
\begin{tabular}{lrrrrrr}
    \toprule
    & \texttt{AE}  & \texttt{VAE} & \texttt{BeatGAN} & \texttt{BeatGAN}$_+$ & \texttt{\nmodel$_{1}$} & \texttt{\nmodel$_{2}$} \\
    \midrule
    TSTR &  $0.869\pm0.017$ &  $0.738\pm0.025$  & $0.881\pm0.138$  & $\mathbf{0.929\pm0.009}$ & $0.881\pm0.012$&  $0.872\pm0.011$ \\
    TSTR$_N$ & $0.885\pm0.017$ & $0.754\pm0.025$ & $0.900\pm0.140$ & $\mathbf{0.951\pm0.015}$ & $0.907\pm0.007$ & $0.891\pm0.011$ \\
    \bottomrule
\end{tabular}
}
\label{tab:tstr}
\end{table*}
\paragraph{Latent Anomaly Score}
\begin{figure*}[!t]
\vspace*{-0.5cm}
\centering
\scalebox{0.420}{\input{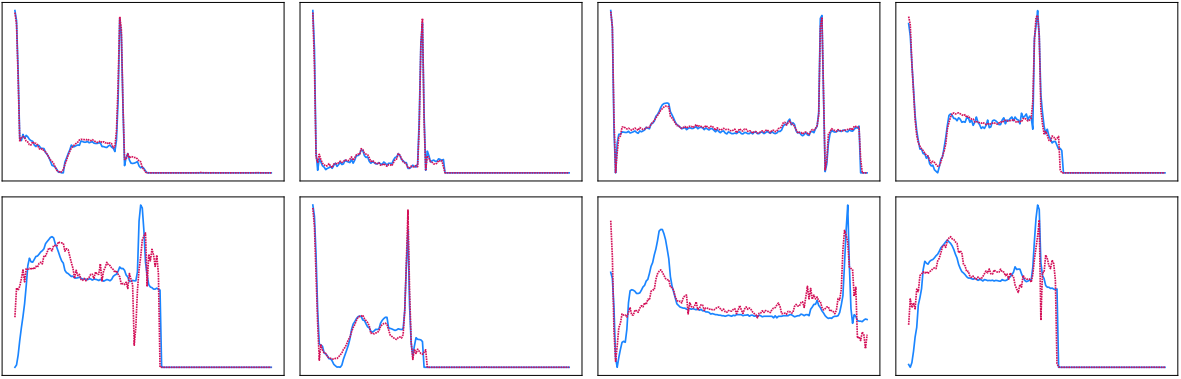}}
\caption{Randomly selected real (blue) and reconstructed (red) samples of the MITBIH dataset. Top: Normal class, bottom: Abnormal class. }
\label{fig:fixed_samples}
\end{figure*}
We can use the information about the $\chi$ distribution to define a new loss $\mathcal{L}_{z}$.
For traditional GANs where $z\sim \mathcal{N}(0,\text{I})$, we can simply use the distance from the norm of the latent vector to determine if the latent vector lies in an abnormal region using
\begin{equation}
\mathcal{L}_{z}=\Vert z \Vert_2-\sqrt{dim(z)-1}.
\end{equation}
This cannot be assumed for (variational) AEs per se since $\mathcal{L}_{\mathbb{KL}}>0$. Two practical solutions come to mind: The mode can be explicitly estimated, or implicitly capture the anomalousness using a nonlinear composition of the anomaly score. In the case of the implicit capture, $\mathcal{L}_{z}=\vert \vert z \vert\vert_2$ can be used.
If the explicit estimation using the posterior distribution is utilized, we can define $\mathfrak{L}$ as the set of all latent norms $\Vert z_i \Vert_2 = \Vert q_\theta(z_i|x_i) \Vert_2, i\in\{1,\dots,m\}$ in the training dataset with $m$ samples and the resulting latent error of a given latent representation $z$ as
\begin{equation}
\mathcal{L}_{z} =\Vert z \Vert_2-\text{mode} ( \mathfrak{L}).
\end{equation}
In case of explicit mode estimation, we bin the latent norms of normal class training data and select the bin with the highest frequency as our mode.
This leads to another trade-off of $\beta$:
A high $\beta$ makes it easier to reliably detect OOD regimes of the latent space. But sufficient reconstructions require a low $\beta$ and assimilation towards the prior can lead to a posterior collapse, where the generated samples lack the required variability.
Additionally to the error components from Schlegl et al. \cite{anogan} we thus introduce the latent loss $L(x)= \mathcal{L}_{\hat{z}}$ resulting in an updated anomaly score 
\begin{equation}
\label{eq:updated_anogan}
A^\ast(x) = (1-\lambda-\gamma) \cdot R(x) + \lambda \cdot D(x) + \gamma \cdot L(x),
\text{with }
\lambda+\gamma\leq 1.
\end{equation}
\paragraph{Criterion Weights}
Lastly, the selection of the weighting parameters needs to be addressed.
Existing algorithms have either used only $R(x)$, only $D(x)$, or an empirically selected $\lambda$ for the linear combination to calculate A(x). Since AD performance is the main objective of our models, it is furthermore useful to investigate AD performance during validation. However, a grid search to optimize $\lambda$ (and $\gamma$) is costly, depending on the selected search space. Additionally, no nonlinear relationships between the error components are considered, even though this  might be necessary in edge cases. Training a kernelized nonlinear SVM on the individual loss components can help solve these problems: Depending on the size of the dataset, the computational costs are reduced significantly and a more robust separating hyperplane can be learned through margin maximization. This makes them especially useful for medical datasets, which usually encompass only several hundred or thousand samples. 
\section{Experiments}
We evaluate \model on the public MITBIH arrhythmia dataset \cite{mitbih} and the CMU Wafer dataset \cite{wafer}. Our baselines include an AE, VAE, the original BeatGAN, and an improved version of BeatGAN, \texttt{BeatGAN}$_+$, which utilizes a reduced latent size and does not apply input normalization in the discriminator but instead uses spectral normalization. These settings are also applied to \nmodel. We compare two different $\beta$: $\beta=0$ for \texttt{\nmodel$_{1}$} and $\beta=0.0001$ for \texttt{\nmodel$_{2}$}. For $\beta=0$ the $\mathbb{KL}$ divergence is ignored and the latent space distribution does not converge towards $\mathcal{N}(0,\text{I})$. We normalize the reconstruction loss using the sequence length, number of channels, and batch size. On the MITBIH dataset the normalization by $\texttt{sequence length}\times \texttt{number of channels} \times \texttt{batch size}=160\times1\times256=40960$, $\beta=0.0001$ corresponds to an unscaled $\beta=40960\times0.0001=4.096$, which approximately equals $\beta=4$ recommended in the original $\beta$-VAE \cite{betavae}. More details on preprocessing, hyperparameters and model settings can be found in the supplementary material.
\subsection{Generative Capabilities}
\addtolength{\tabcolsep}{-1.5pt}
\begin{table*}[!b]
\centering
\caption{Weighted $F_1$ score of anomaly score compositions averaged across five folds.}
\resizebox{\linewidth}{!}{%
\begin{tabular}{lrrrr|rrrr}
    \toprule
    & \multicolumn{4}{c}{MITBIH} &  \multicolumn{4}{c}{Wafer}\\
    \cmidrule(lr){2-5} \cmidrule(lr){6-9}
    & \texttt{BeatGAN} &  \texttt{BeatGAN}$_+$ & \texttt{\nmodel$_{1}$} & \texttt{\nmodel$_{2}$} & \texttt{BeatGAN} &   \texttt{BeatGAN}$_+$ & \texttt{\nmodel$_{1}$} & \texttt{\nmodel$_{2}$}  \\
    \midrule
    $\lambda=0$  &  $0.854\pm0.002$ & $0.899\pm0.002$  & $0.905\pm0.003$ & $0.909\pm0.002$&
    $\mathbf{0.727\pm0.082}$ & $0.972\pm0.011$&$0.980\pm0.006$&$0.983\pm0.003$ \\
    
    $\lambda_{GRID}$  & $0.853\pm0.002$ & $0.899\pm0.003$ & $0.905\pm0.003$ & $0.909\pm0.002$ &
    $0.709\pm0.070$&$0.980\pm0.005$&$\mathbf{0.980\pm0.005}$&$0.984\pm0.008$ \\
    
    $SVM_\lambda$  & $0.856\pm0.004$ & $0.906\pm0.003$ & $\mathbf{0.915\pm0.002}$ & $\underline{\mathbf{0.920\pm0.003}}$ &
    $0.570\pm0.109$&$0.980\pm0.003$&$0.979\pm0.009$&$0.984\pm0.005$ \\
    
    $SVM_\gamma$  & $\mathbf{0.878\pm0.005}$ & $\mathbf{0.908\pm0.002}$ & $0.914\pm0.003$ & $0.919\pm0.004$&
    $0.715\pm0.071$&$\mathbf{0.982\pm0.004}$&$0.980\pm0.007$&$\underline{\mathbf{0.986\pm0.008}}$ \\
    \bottomrule
\end{tabular}
}
\label{table:modelselection}
\end{table*}
All models are trained across 500 epochs using 5-fold cross-validation  
using a train set containing only normal class data. Validation and test data consist of normal as well as abnormal data. 
Normal samples need to be reconstructable, while abnormal samples should not be reconstructable. Random samples (Fig. \ref{fig:fixed_samples}) show that this is most often the case, but some abnormal data is still reconstructable to a high degree, highlighting a drawback of AD solely relying on reconstructions. 
We find that Maximum Mean Discrepancy is not very meaningful to quantify the similarity of real and synthetic data if \tG is an AE, quickly converging towards 0 in our experiments.
We thus utilize TSTR \cite{rgan} and additionally introduce the normalized TSTR score, $\text{TSTR}_N=\frac{F_1(\text{classifier}_{synth})}{F_1(\text{classifier}_{real})},$ to account for the general complexity of the dataset. Results using CNN classifiers are reported in Table \ref{tab:tstr}. 
TSTR is performed on the entire validation dataset and is reshuffled in each epoch. To account for the fluctuation in single TSTR scores introduced by the reshuffling, we use the moving average across 5 measurements. All models achieve reasonably high TSTR scores, with \texttt{\nmodel$_2$} achieving lower TSTR scores than the (modified) BeatGAN architecture. This is to be expected since the higher $\beta$ of \texttt{\nmodel}$_2$ restricts the generative quality. 
\subsection{Anomaly Detection}
\begin{figure}[!h]
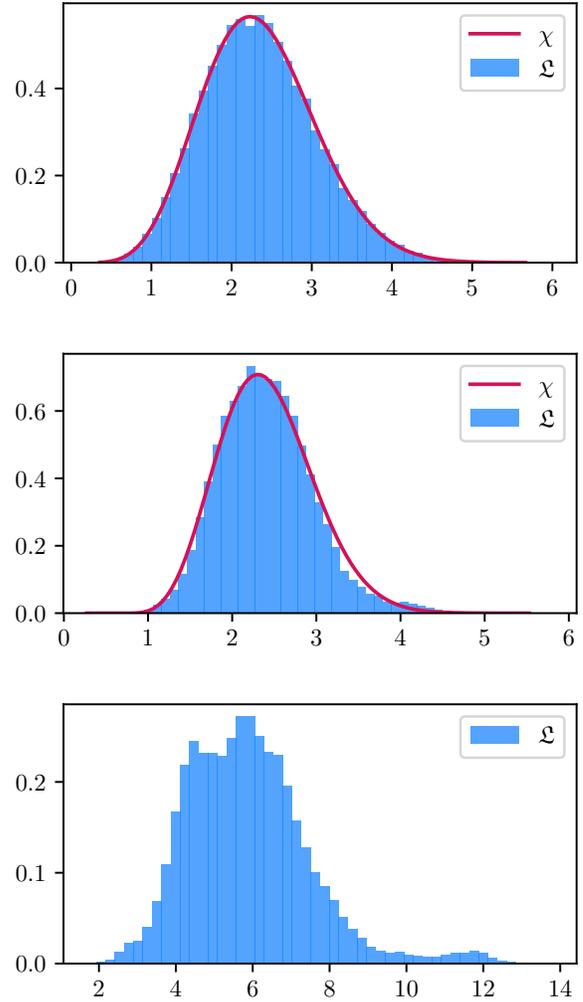

\centering
\resizebox{0.45\textwidth}{!}{\input{chi_beta_dot1_1.pgf}}
\resizebox{0.45\textwidth}{!}{\input{chi_beta_dot0001.pgf}}
\resizebox{0.45\textwidth}{!}{\input{chi_beta_dot0.pgf}}
\caption{$\chi$ distribution fit on latent norms of the MITBIH training data using \model with a latent size of 6. Comparison of  $\beta=0.1$ (top), $\beta=0.0001$ (mid) and $\beta=0.0$ (bottom).The $\chi$ distribution fit of $\beta=0.1$ with a KS statistic of 0.0029 and p = 0.89 indicates high agreement for $\mathfrak{L}\sim \chi$. No sufficient $\chi$ fit was possible on $\beta=0.0$.}
\label{fig:chi}
\end{figure}

\begin{figure}[ht]
\centering
\resizebox{0.45\textwidth}{!}{\input{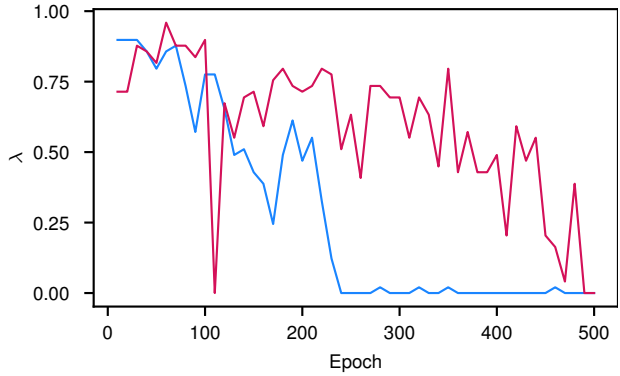}}
\caption{Progression of the optimal $\lambda$ across two \model folds on the MITBIH dataset on validation data.}
\label{fig:progression}
\end{figure}

\paragraph{Optimal $\lambda$} We investigate the optimal $\lambda$ from the linear anomaly score composition from A(x) using a grid search during validation. $\lambda$ is non-stationary even though the reconstruction error is usually favored as training progresses in our experiments. $\lambda>0$ often remains beneficial and the optimal $\lambda$ can differ significantly when hyperparameters are changed. The progression of $\lambda$ during two different folds has been visualized in Fig. \ref{fig:progression}. It can be seen that the optimal composition of the anomaly score components changes significantly over time.
\paragraph{$\chi$ distribution} To validate the theoretical motivation from Section \ref{section:ad} we train \model with $\beta = 0.1$ (effectively an unnormalized $\beta=4096$). We show high agreement for the hypothesis of the latent norms following the $\chi$ distribution, cf. Fig. \ref{fig:chi}. However, the $\mathbb{KL}$ divergence is favored too strongly if $\beta=0.1$, leading to a posterior collapse. A lower $\beta$ is required for AD. Comparisons of the distributions of the latent norms of training data with various $\beta$ can be found in the supplementary material.
\paragraph{Anomaly score composition} We compare the anomaly score of AnoGANs' $A(x)$ with grid searched $\lambda_{GRID}$ to using only the reconstruction error R(x) $(\lambda=0)$, an RBF-SVM optimized score using $R(x)$ and $D(x)$ in $SVM_\lambda$ and additionally add the latent score $L(x)$ to $SVM_\gamma$, also using a SVM with RBF kernel. 
$\lambda_{GRID}$ as well as the SVMs are trained on validation data and evaluated on test data. $\lambda_{GRID}$ and the SVMs both are trained to maximize the $F_1$ score. Each error is min-max normalized to limit the search space of the grid search. Results are available in 
Table \ref{table:modelselection}. $SVM_\gamma$ and $SVM_\lambda$ perform best for the models investigated, including \texttt{BeatGAN}. \nmodel$_2$ achieves the highest performance on both datasets.
\addtolength{\tabcolsep}{3.3pt} 

\paragraph{Comparison to baselines} 
\begin{table*}[b]
\centering
\caption{Comparison of AD performances averaged across five folds on test data with parameters from validation data.}
\resizebox{\linewidth}{!}{%
\begin{tabular}{lccc|ccc}
   \toprule
    & \multicolumn{3}{c}{\textbf{MITBIH}} &  \multicolumn{3}{c}{\textbf{\secondDataset}}\\
    \cmidrule(lr){2-4}
    \cmidrule(lr){5-7}
     & $F_1$     & $r_\phi$     & $AUROC$       & $F_1$ & $r_\phi$     & $AUROC$   \\
     \midrule
\texttt{AE} &  
$0.902\pm0.003$&   
$0.796\pm0.005$&   
$0.900\pm0.004$&       
    $0.978\pm0.004$&           
    $0.951\pm0.006$&           
    $0.982\pm0.003$
       \\
\texttt{VAE}                     &       
$0.883\pm0.006$&           
$0.757\pm0.011$& 
$0.883\pm0.006$&       
$0.957\pm0.012$&           
$0.905\pm0.030$&           
$0.964\pm0.011$
\\
\texttt{BeatGAN}    & 
$0.878\pm0.005$    &    
$0.746\pm0.009$        &   
$0.862\pm0.008$&       
$0.727\pm0.082$&           
$0.378\pm0.190$&           
$0.613\pm0.091$\\
\texttt{BeatGAN$_+$}    &      
$0.908\pm0.002$    &    
$0.808\pm0.005$        &   
$0.897\pm0.003$  &       
$0.982\pm0.004$&           
$0.956\pm0.011$&           
$\mathbf{0.984\pm0.002}$   \\
\texttt{\nmodel$_{1}$}  &     
$0.915\pm0.002$    &    
$0.822\pm0.003$       &   
$0.908\pm0.002$    &       
$0.980\pm0.005$&           
$0.950\pm0.012$&           
$0.978\pm0.009$  \\
\texttt{\nmodel$_{2}$}  &     
$\mathbf{0.920\pm0.003}$    &    
$\mathbf{0.832\pm0.006}$       &   
$\mathbf{0.918\pm0.004}$    &       
$\mathbf{0.986\pm0.008}$&          
$\mathbf{0.964\pm0.020}$&           
$0.980\pm0.013$ \\
\bottomrule
\end{tabular}
}
\label{table:performance}
\end{table*}
\addtolength{\tabcolsep}{1.5pt} 
The AD performance of all models is compared in Table \ref{table:performance}, including the $F_1$ score, phi coefficient $r_\phi$ and AUROC of the best-performing methods from Table \ref{table:modelselection}. The reconstruction error is used for AE and VAE. \texttt{\nmodel} achieves the best performances  on the MITBIH dataset.
\paragraph{Generalization}
We further evaluate the model on the CMU Wafer dataset. It is especially of interest because of the discrete levels of the measurements, which is difficult given common problems with the noisy reconstructions of AEs. The hyperparameters are used as in the MITBIH experiments with the only difference being a reduced latent dimensionality of 4.
While being noisy, the reconstructions still allow a high (reconstruction-based) AD performance due to the model's inability to reconstruct abnormal data.
Results can be found in Table \ref{table:modelselection} and Table \ref{table:performance}. Table \ref{table:performance} shows a drawback of solely using the $F_1$ score for AD if the dataset is imbalanced: \texttt{BeatGAN} achieves reasonably high $F_1$ scores without learning sufficient reconstructions, visible in $r_\phi$ and AUROC. Further experimental insights are available in the supplementary material.
\section{Discussion and Future Work}
\paragraph{Data generation}
We find that a high latent dimensionality and resulting little compression is one of the main reasons of the poor performance of \texttt{BeatGAN} on the MITBIH dataset. The second major modification is the use of spectral normalization in the discriminator, leading to the most significant improvements in our experiments. Spectral normalization especially helps to improve the discrimination error but also improves the reconstructive capabilities due to the adversarial loss component in $\mathcal{L}_\G$. 
We only perform limited experiments with different hyperparameters, extensive ablations are likely to further improve generation as well as anomaly detection.

\paragraph{Anomaly detection}
$\lambda$, as well as $\gamma$, can reduce robustness if the model is not sufficiently trained. The results in Table \ref{table:modelselection} show that for individual models, each score composition can be the most robust. One reason is that we utilize the parameters of the highest-performing validation epoch, which sometimes occurs early on during training and leads to less robust predictions on the test data set. Comparing the selection of $\lambda$ using the grid search as well as SVM, the SVM-based models most often improve robustness, which is to be expected given the margin maximization. The performance itself is enhanced by using the SVM, most notably improving the $F_1$ score from 0.909 to 0.920 in the best performing \nmodel$_2$.
The time complexity of training an SVM does not necessarily allow the use on significantly larger datasets, cf. supplementary materials.\\
While the use of the latent norm improves the AD in most experiments and SVM$_\gamma$ achieved the highest performance in individual models of both datasets, the influence of $L(x)$ is most often small in comparison to $D(x)$ and $R(x)$. The lower impact is to be expected since \tG should not be able to generate the samples at all and $\gamma$ only becomes relevant if the model generates samples far from the mean of the latent space. The results indicate that the generation of such points is unlikely and the first experiments show that the importance of the latent norm gets more relevant if manipulations of the latent space are performed as in the AnoGAN interpolations. Both $L(x)$ as well as the total (euclidean) distance traversed through latent space have been indicators of abnormal behavior in the AnoGAN optimization scheme. Apart from this, it is furthermore desirable to use traditional GANs with a predefined $p_z$ where the $\mathbb{KL}$ divergence does not influence the reconstruction quality. However, the resulting generated samples using various traditional GANs did not yield similar quality in our experiments.\\ 
Interestingly, the latent norm loss component has an impact on the BeatGAN AD performance, improving the $F_1$ score from 0.856 (SVM$_\lambda$) to 0.878 (SVM$\gamma$) with similar robustness. This is an unexpected result since the latent norm of traditional AEs is not expected to follow any specific distribution. We find that the distribution of latent norms of deterministic AEs is consistently unimodal in the training data in the early stages of the training and bimodal on validation data which includes abnormal samples, cf. suppl. material. The shape of the distribution as well as the range of values of this distribution change over time, requiring future investigations. Apart from using the $\chi$ distribution assumption, possible restrictions include stationary latent norm limitations of the AnoGAN interpolation or the use of truncated Gaussians. BigGAN\cite{biggan} utilizes truncated Gaussians to control variety (large truncation range) and fidelity (narrow truncation range) of generated samples which might be useful for GAN-based AD.\\
\model achieves the best AD performance in our experiments and additionally exhibits desirable latent space properties such as less extreme latent values and no significant shift to higher mean latent norms. This compact latent space is even present with $\beta =0$, also requiring further investigations.\\
Apart from significantly improving the reconstructions, we are also able to reduce the inference time from several seconds per sample to below 2ms, similar to the performance of BeatGAN. First findings on interpolating after performing the inverse mapping in an attempt to improve the anomaly detection performance can be found in \ref{appendix:inference}. A promising approach is to transfer this methodology to diffusion models which have recently been used for anomaly detection, e.g. \cite{diffusion}.
\section{Conclusion}
We improve state-of-the-art one-class data generation and anomaly detection for time series data. We advance existing work on adversarial AEs for anomaly detection, e.g. applying spectral normalization to the discriminator. We propose \nmodel, extending existing work by integrating the probabilistic $\beta$-VAE framework into generative anomaly detection. By combining properties of the $\chi$ distribution with the posterior estimation of VAEs, we introduce a novel measure to detect abnormal data based on the norm of the latent vector. While motivated by standard Gaussian latent spaces, we find that the component also improves anomaly detection in traditional adversarial AEs and introduces deviations of the latent norms from the $\chi$ distribution as a way to measure deviations from normal class samples.
We further show that anomaly score compositions change over time and a continuous evaluation of the scoring is required. To achieve this, we optimize the anomaly score components using SVMs with nonlinear kernels, allowing faster, more robust, and better anomaly detection performances.
\section*{Acknowledgments}
This work was supported by the Bavarian Research Foundation under grant AZ-1419-20.
\bibliographystyle{IEEEtran}
\bibliography{IEEEabrv,sample-base}
\clearpage
\appendix
\section*{Reproducibility}
All the code required to reproduce the experiments can be found at \url{https://github.com/emundo/ecgan}. 
It includes instructions on the reproduction, dataset details and functions to automatically download the data. The complete configurations and source code on training runs/experimental results as well as the data and scipts to reproduce the figures presented in this work is also available. 
\section*{Architectural details}
\label{appendix:architecture}
The architecture of our $\beta$-VAE generator consists of an encoder with five layers of 1D convolutional blocks, batch normalization, and Leaky ReLU (leakage: 0.2) before estimating $\mu, \sigma$ of the posterior distribution. As gradients cannot flow through stochastic nodes, the reparameterization $z= \mu_\phi+\sigma_\phi \cdot \epsilon$ with $\epsilon \sim \mathcal{N}(0,\text{I})$ is required for end-to-end optimization. The decoder consists of five layers of transposed 1D convolutions, each followed by batch normalization and ReLU activations before returning $\hat{x}$ using a sigmoid layer. 
The discriminator consists of five layers with 1D convolution blocks, followed by spectral weight normalization \cite{spectral_norm} and Leaky ReLU activations (leakage: 0.2) before retrieving the scalar discriminator output from another sigmoid layer. 
\subsection{Spectral normalization}
\begin{figure}[H]
  \centering
    \setlength{\abovecaptionskip}{-5pt} 
    \subfloat[$F_1$ score of using only $D(x)$ as anomaly indicator comparing BeatGAN with latent size 6 with and without spectral normalization.]{%
      \scalebox{1.0}{\input{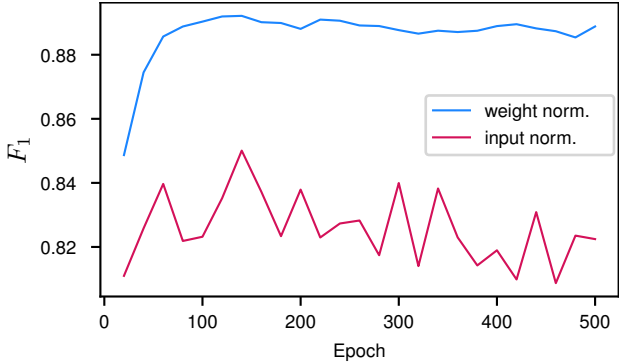}}
            \label{fig:spectral_norm}
      }
      \qquad
    \subfloat[$F_1$ score of SVM$_\gamma$ comparing BeatGAN with latent size 6  with and without spectral normalization.]{%
      \scalebox{1.0}{\input{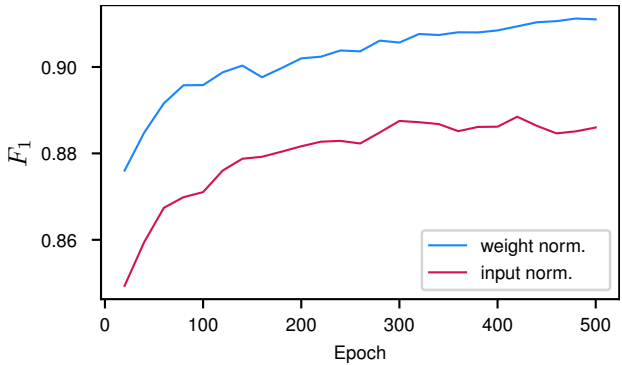}}\label{fig:spectral_norm_combined}}
      \vspace*{0.5cm}
          \caption{Spectral normalization.}
\label{fig:sn}
\end{figure}
\vspace*{-0.5cm}
\subsection{Maximum latent values}
\begin{figure}[h]
\centering
\resizebox{0.48\textwidth}{!}{\input{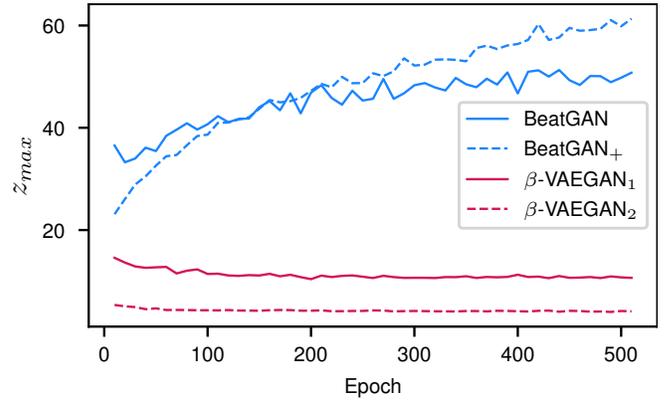}}
\caption{Maximum values of the latent vectors across the training set comparing BeatGAN and \model. \model produces similar maximal and minimal values which are stable over time while while the maximum values drift apart for BeatGAN and BeatGAN$_+$}
\label{fig:max_val}
\end{figure}
\section*{Experimental setup}
\label{appendix:experiments}
We use beat-wise preprocessed data from Kachuee et al. \cite{mitbih_beats}. No samples are removed and we utilize the first channel of all patients.
To investigate generality, we additionally investigate the CMU \secondDataset dataset, containing sensor data to detect abnormal silicon wafers during manufacturing.\\
The original BeatGAN serves as our first baseline. 
Furthermore, we develop an improved version of BeatGAN, \texttt{BeatGAN}$_+$.
\texttt{BeatGAN}$_+$ does not apply input normalization in the discriminator but instead uses spectral normalization. We execute 3 discriminator updates per generator update, the Adam optimizer is replaced by the AdaBelief optimizer with $LR=0.0002$, $\beta_1=0.5$, $\beta_2=0.99$ and $\epsilon=1\times 10^{-16}$ and the tanh output layer of BeatGAN is replaced by a sigmoidal layer in \texttt{BeatGAN}$_+$. We further reduce the latent size to 6 in \texttt{BeatGAN}$_+$ since the compression of BeatGANs' $enc_\phi$ is very limited using the original latent dimensionality of 50. These settings are also applied to \nmodel. We normalize the reconstruction loss by $(\texttt{sequence length}\times \texttt{number of channels} \times \texttt{batch size})$, easing comparisons of $\beta$ across experiments. We compare two different $\beta$: $\beta=0$ for \texttt{\nmodel$_{1}$} and $\beta=0.0001$ for \texttt{\nmodel$_{2}$}. For $\beta=0$ the $\mathbb{KL}$ divergence is ignored and the latent space distribution does not converge towards $\mathcal{N}(0,\text{I})$. On the MITBIH dataset the normalization by $\texttt{sequence length}\times \texttt{number of channels} \times \texttt{batch size}=160\times1\times256=40960$, $\beta=0.0001$ corresponds to an unscaled $\beta=40960\times0.0001=4.096$, which approximately equals $\beta=4$ recommended in the original $\beta$-VAE \cite{betavae}. We use a batch size of 256 for all models except for \texttt{BeatGAN}, where we use the original configuration with a batch size of 64. \\
During training we utilize 85\% normal data, the remaining 15\% are equally split and used during validation and testing using stratified cross-validation. This results in approximately balanced validation/test datasets with initial imbalances of 17.2\% and 10.6\% of abnormal data for the MITBIH and Wafer datasets respectively. The data splits are saved as index lists and can be used to reconstruct train/test/validation datasets. The sequence length of both datasets is slightly changed: To allow compatibility with BeatGAN we utilize a sequence length of 160. For MITBIH the data is reduced from a sequence length of 187 using Largest Triangle Three Buckets\footnote{\url{https://github.com/sveinn-steinarsson/flot-downsample/}} and upsampled from 152 using linear interpolation. Both resampling techniques do not lead to significant visual changes or changes in the performance of baseline classifiers.
Configuration files for training as well as anomaly detection are part of the supplementary code. Tracking is performed using Weights and Biases.
\subsection{AE/VAE Baselines}
Our baselines include the reconstruction error of an AE as well as a VAE. Both utilize the same optimizer settings (AdaBelief optimizer with $LR=0.0002$, $\beta_1=0.5$, $\beta_2=0.99$ and $\epsilon=1\times 10^{-16}$) and latent size (6) as \nmodel.
We utilize the same architecture as in the generators of BeatGAN/\nmodel, and perform a grid search across several configurations. These especially include using the MSE if using a \textit{tanh} output function and BCE error if using a sigmoidal output layer. We find that the sigmoidal output with BCE performs best in our experiments. Additionally, we use various $\beta$. We find that $\beta=0.0001$ from \model without the adversarial components does not allow sufficient reconstructions for the VAE and utilize the best performing $\beta=0.0$ in our experiments .\\
First experiments on changes of $\mathcal{L}_\D$ to also consider $\mathcal{L}_{adv}$ as a weighted component or as the sole component did not yield any advantages. We only consider $\nu=1$  \eqref{eq:gen_loss}.
Additionally, the use of an OCSVM has been investigated. Even after reducing the dimensionality to 50 using PCA and applying a grid search for the OCSVMs hyperparameters, this resulted in a weighted $F_1<0.3$. We thus abstain from reporting the full results in the main body of this work, the implementation of the OCSVM is available in the supplementary code. 
\section*{Supplementary experimental results}
\label{appendix:experimental_results}
\subsection{TSTR}
\begin{figure}[!h]
\centering
\resizebox{0.48\textwidth}{!}{\input{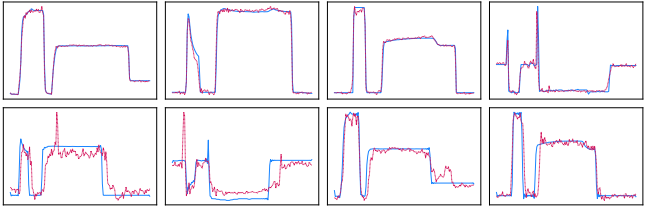}}
\caption{Randomly selected reconstructed samples of the Wafer dataset using \nmodel. The top row consists of normal class data, the bottom row of abnormal data. Abnormal samples are noisy and not reconstructible, normal samples still suffer from small deviations and are not as smooth as the original data.}
\label{fig:fixed_samples_wafer}
\end{figure}

We find that the visual similarity can be deceptive during training: Reconstructions usually are of high quality after a few epochs but TSTR as well as AD performance often still consistently improve after hundreds of epochs. The experiments further show a drawback of TSTR. The TSTR results on the CMU Wafer dataset (Table \ref{tab:tstr_wafer}) indicate a near-perfect reconstruction with TSTR$_N=1$. While the resulting reconstructions on abnormal data are very dissimilar and the normal class data can be reconstructed with a low MSE, the normal class reconstructions exhibit atypical noise, cf. Fig. \ref{fig:fixed_samples_wafer}. One possible explanation is that the underlying CNN classifier identifies abnormal data based on the high frequencies present in the wafer anomalies, ignoring the lower frequent reconstruction noise. The common interpretation and use of TSTR as measuring the similarity between the real and synthetic data can thus be misleading since it mainly measures the \textit{separability} of the real and synthetic normal and abnormal data.\\
\begin{table}[h]
\caption{Wafer TSTR scores across five folds (running mean of 5 measurements).}
\centering
\begin{tabular}{lrr}
\hline
&\textbf{TSTR}&\textbf{TSTR$_N$}
\\
\texttt{AE} & 
$0.994\pm0.001$  &
$0.999\pm0.002$ %
\\
\texttt{VAE} & 
$0.982\pm0.008$  &
$0.988\pm0.006$ %
\\
\texttt{BeatGAN} & 
$0.173\pm0.005$  &
$0.175\pm0.004$ %
\\
\texttt{BeatGAN}$_+$& 
$0.995\pm0.001$&
$\mathbf{1.000\pm0.001}$  
\\
\texttt{\nmodel$_{1}$} & 
$0.995\pm0.001$&
$1.000\pm0.001$  
\\
\texttt{\nmodel$_{2}$}& 
$\mathbf{0.996\pm0.001}$&
$1.000\pm0.001$ 
\\
\hline
\end{tabular}
\label{tab:tstr_wafer}
\end{table}
\subsection{Interpolation Grid}
\begin{figure*}[h]
\centering
\includegraphics[width=\textwidth]{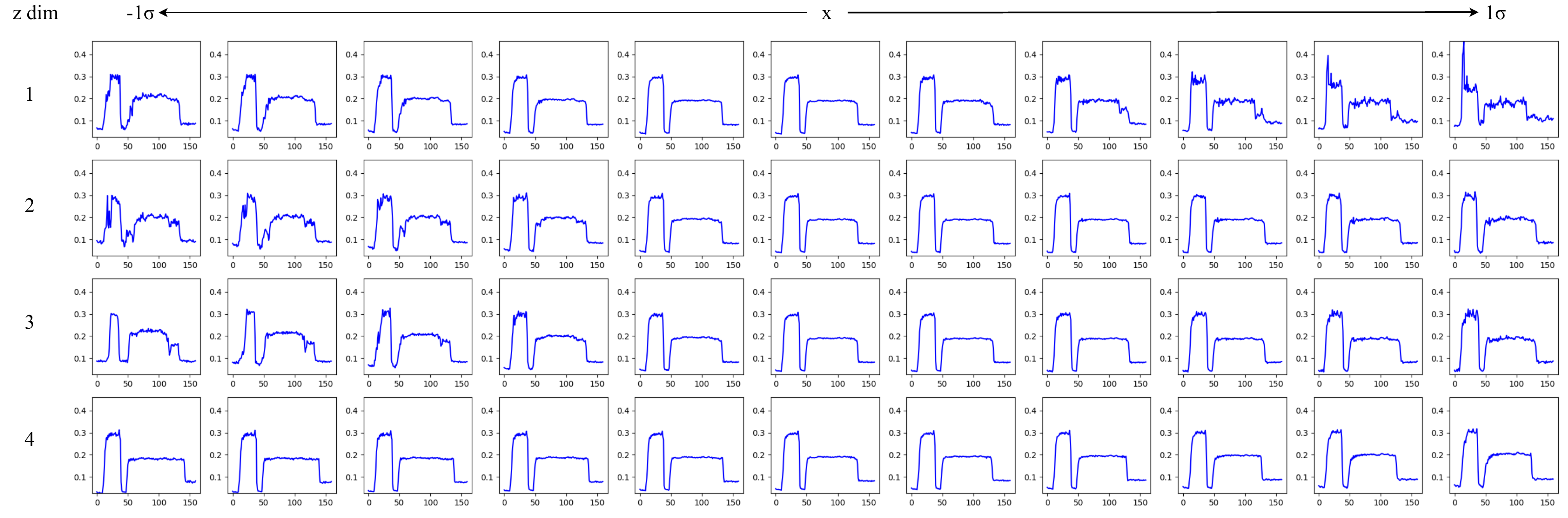}
\caption{Interpolation on the Wafer dataset. Each row is an interpolation starting from the middle column interpolating in one dimension. The first sample of each row corresponds to the sample farthest away from the original sample in the negative direction, the last sample of each row is farthest away in the positive direction.}
\label{fig:wafer_interpolation_grid}
\end{figure*}

Fig. \ref{fig:wafer_interpolation_grid} demonstrates interpolations in latent space into the 4 latent dimensions used in the \nmodel$_2$ experiments. While some dimensions become abnormal with increasing interpolation distance, the encoding of several features is clearly visible.
\subsection{BeatGAN Reconstructions on Wafer Data}
\begin{figure}[!h]
\centering
\resizebox{0.48\textwidth}{!}{\input{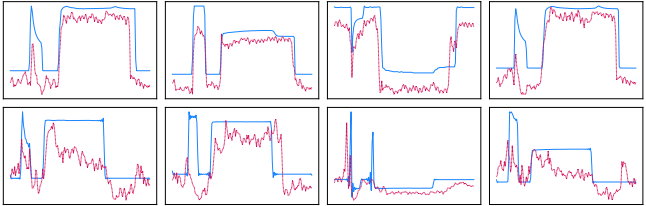}}
\caption{Randomly selected reconstructed samples of the Wafer dataset using the BeatGAN architecture. Top rows consist of normal class data, bottom rows of abnormal data. Neither class can be sufficiently reconstructed with BeatGAN.}
\label{fig:fixed_samples_beatgan_wafer}
\end{figure}

Tables \ref{table:modelselection} and \ref{table:performance} show that the basic \texttt{BeatGAN} setting does not allow for sufficient reconstructions: While the weighted $F_1=0.73$ indicates that the reconstructions can be useful for AD, $r_\phi$ close to 0 and AUROC close to 0.5 indicate that the model is almost randomly guessing the class membership. This is emphasized by high deviations from normal class data and the visual dissimilarity in Fig. \ref{fig:fixed_samples_beatgan_wafer}.
The stark differences between BeatGAN as well as BeatGAN$_+$ and the (V)AE baselines are likely due to the tanh output of the original BeatGAN which was replaced with a sigmoidal output layer in the remaining models, significantly improving the quality. 
The remaining experiments support the improvements in generative as well as AD quality, achieving equal or better quality in the AD performance. 
\subsection{Latent norm distribution}
\begin{figure}[H]
\includegraphics[width=0.45\linewidth]{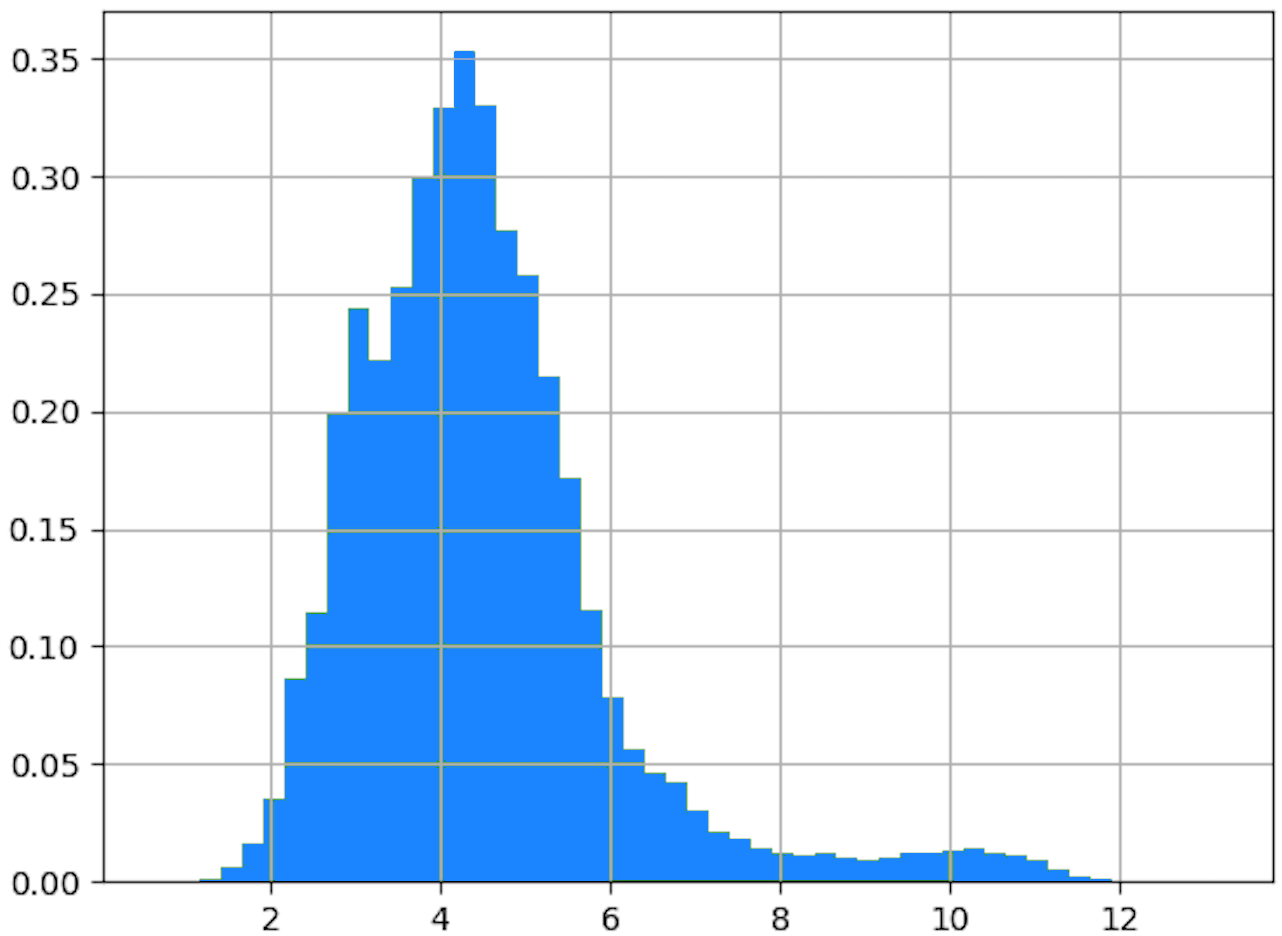} \includegraphics[width=0.45\linewidth]{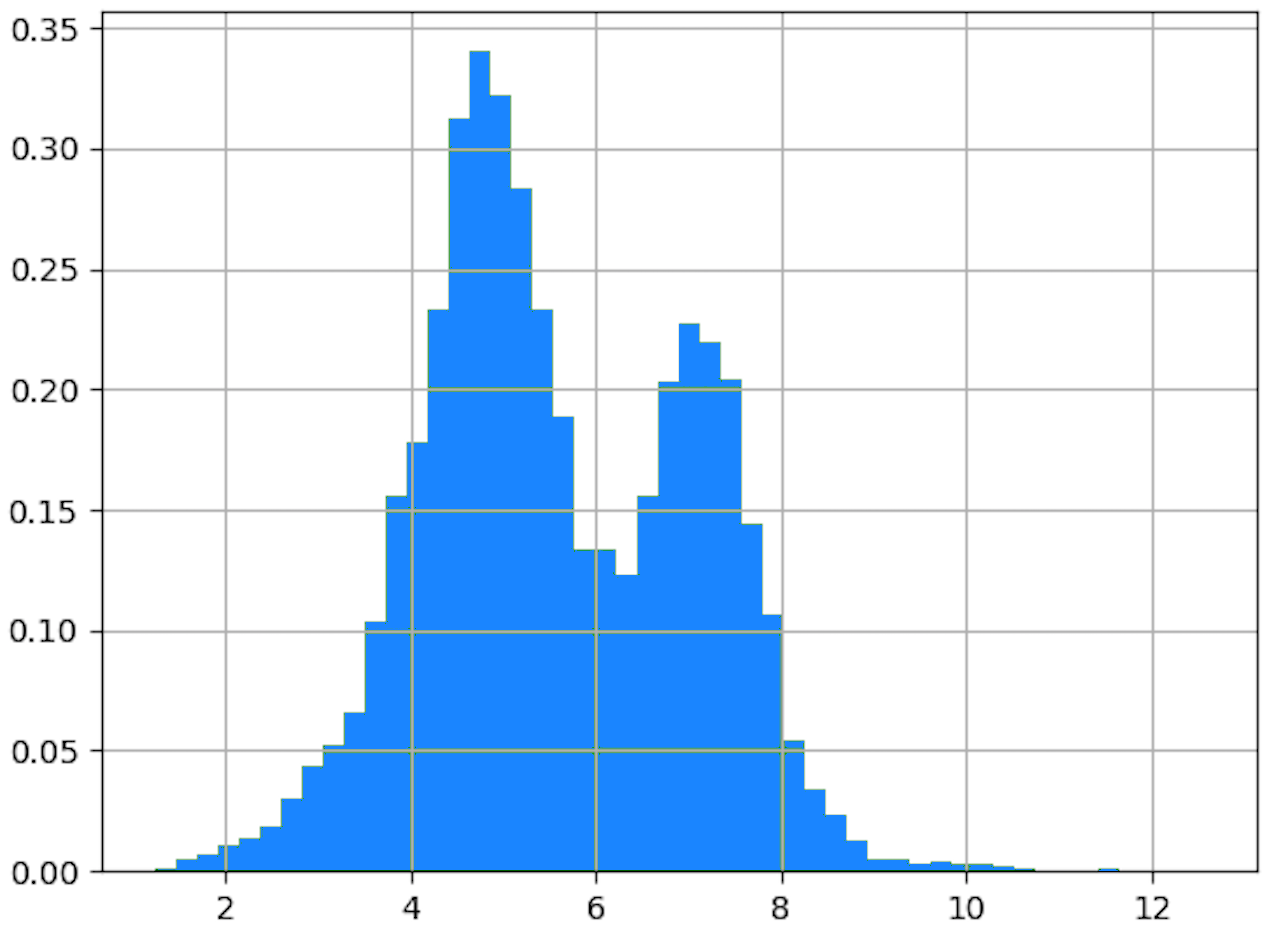}
\caption{Histograms of latent norms from BeatGAN. Left: Train data (only normal class), consistently resulting in a unimodal long-tailed distribution. Right: Validation data (normal as well as abnormal class), resulting in a bimodal distribution.}
\label{fig:latent_norms_hist}
\end{figure}
\subsection{Time reduction to optimize $\lambda$}
\label{appendix:time}
 While the feature dimensionality is generally low (the amount of errors, i.e. 2 or 3 in our experiments), the influence of the amount of samples $n$ is important, with training times usually between $\mathcal{O}(n^2)$ and $\mathcal{O}(n^3)$ for nonlinear SVMs depending on implementation details as well as hyperparameters \cite{svm_runtime}. The grid search can be significantly more costly, depending on the size of the dataset and the size of the grid. On the MITBIH data ($n=109446$) with a grid of 50 possible $\tau$ and 100 $\lambda$ values, we are able to reduce the time required to find $\lambda$ from 43s using a grid search to 7s using SVM$_\lambda$ on a 2,6 GHz 6-Core CPU. It is furthermore difficult to tune the search range of the grid search, easily leading to low robustness (thin grid) or bad initial results (coarse grid).
 \subsection{Inference}
 \label{appendix:inference}
 Using the AnoGAN latent optimization we require between 5s and 30s for a single sample on the previously described CPU, depending on the desired similarity $\epsilon$ and the maximum amount of interpolation steps. For BeatGAN and \model the inference time is less than 2 ms, confirming the results in Zhou et al. \cite{beatgan}. Additionally we investigate interpolation after retrieving the inverse mapping. This often results in significantly improved reconstructions, also of anomalous data. The high variance of inference times ($0.48 \pm 1.54$s), is especially influenced by samples which cannot be reconstructed to $\epsilon$-similarity. Retraining the AD parameters using this combined procedure during validation increases the time required such that its evaluation is unfeasible on customer GPUs, making it generally undesired for validation during training. Using the original anomaly score compositions from the inverse mapping of the encoder, we find that the AD performance does not significantly differ and frequently slightly degrades. This indicates the previously shown strong dependence on the reconstruction error: The reduced similarity in input space cannot be compensated by the discriminative ability of our model. Further investigations are of interest if the discrimination score is dominant.
\end{document}